# Oceanship: A Large-Scale Dataset for Underwater Audio Target Recognition

Zeyu Li[1,2][0009-0009-1923-5989], Suncheng Xiang[1,3], Tong Yu[1,2], Jingsheng Gao[1,2], Jiacheng Ruan[1,2], Yanping Hu[1,2], Ting Liu[1,2], and Yuzhuo Fu[1,2,*]

[1] Shanghai Jiao Tong University, Shanghai Minhang 200240, China
{lizeyujack, xiangsuncheng17, tyad2021, gaojingsheng, jackchenruan, yanping_hu, louisa_liu, yzfu}@sjtu.edu.cn
[2] School of Electronic Information and Electrical Engineering
[3] School of Biomedical Engineering

**Abstract.** The recognition of underwater audio plays a significant role in identifying a vessel while it is in motion. Underwater target recognition tasks have a wide range of applications in areas such as marine environmental protection, detection of ship radiated noise, underwater noise control, and coastal vessel dispatch. The traditional UATR task involves training a network to extract features from audio data and predict the vessel type. The current UATR dataset exhibits shortcomings in both duration and sample quantity. In this paper, we propose Oceanship, a large-scale and diverse underwater audio dataset. This dataset comprises 15 categories, spans a total duration of 121 hours, and includes comprehensive annotation information such as coordinates, velocity, vessel types, and timestamps. We compiled the dataset by crawling and organizing original communication data from the Ocean Communication Network (ONC) database between 2021 and 2022. While audio retrieval tasks are well-established in general audio classification, they have not been explored in the context of underwater audio recognition. Leveraging the Oceanship dataset, we introduce a baseline model named Oceannet for underwater audio retrieval. This model achieves a recall at 1 (R@1) accuracy of 67.11% and a recall at 5 (R@5) accuracy of 99.13% on the Deepship dataset.

## 1 Introduction

Underwater Acoustic Target Recognition (UATR) is vital in the field of vessel acoustics as it aims to automatically identify and analyze the emitted sound from targets [1]. This identification of vessels through their sound offers valuable insights into the origins of noise in underwater environmental monitoring systems. During the early stages of UATR (Underwater Acoustic Target Recognition) development, a significant amount of work was conducted privately due to the limited availability of datasets. Researchers employed audio feature extractors, including MFCC [2], GFCC [3], CQT [4], to analyze underwater audio signals [5]. However, in the past decade, advancements in deep neural networks and the introduction of two specific datasets, Deepship [4] and Shipsear [6], have allowed researchers to shift their focus towards preprocessing underwater audio and training neural networks for supervised classification tasks [7].

In the domain of underwater image recognition, Liu et al. [8] initially emphasized the generalization capability of underwater target recognition. However, no research has yet investigated the generalization ability of models in the UATR task. Drawing inspiration from the CLAP task [9], we can evaluate the generalization ability of UATR models through zero-shot classification and retrieval tasks. However, due to the limited availability of UATR data, conducting experiments with zero-shot and retrieval task settings alone is insufficient. To bridge this gap, we propose Oceanship, a dataset for UATR that is characterized by its descriptive richness. It comprises 107,540 audio-text pairs. Moreover, Oceanship surpasses existing datasets by over 230 times in terms of scale, making it an ideal choice as a pretraining dataset for the UATR generalization task. Oceanship encompasses the entirety of the categories found in Deepship, while it encompasses a significant portion, approximately 50%, of the categories found in Shipsear. This extensive coverage simplifies auxiliary procedures within the cross-domain model, facilitating domain adaptation and generalization. Notably, the European Union has already implemented data privacy regulations, such as the General Data Protection Regulations (GDPR) [10], to safeguard sensitive information. In our dataset, we meticulously convert MMSI information into IDs to ensure that private vessel details remain undisclosed. The process of acquiring the data is illustrated in **Fig. 1**, with further details provided in Section 2.

The state-of-the-art (SOTA) UATR model, UART [7], stands out as the pioneering approach to integrating multimodal techniques into the field of UATR. UART achieves training by aligning features extracted from wavelet spectrogram, text, and audio modalities. In comparison to unimodal models, UART exhibits superior recognition capability and enhanced generalization performance. In the realm of general classification tasks, CLIP [11] introduced the concept of large-scale pretraining to achieve text-image alignment, thereby holding great relevance for cross-modal retrieval. Building upon this foundation, AudioCLIP[12] made significant strides by indirectly aligning the audio and text modalities, representing the pioneering effort in cross-modal alignment within the broader field of general audio classification. Taking a step further, ImageBind [13] leverages a unified embedding approach to encompass six distinct modalities, including images, text, audio, depth, thermal, and IMU data. It is important to note that the references have been updated accordingly.

With the emergence of large-scale models like ChatGPT[1] and Gemini[2], the general training process has experienced a gradual shift towards fine-tuning on pre-trained models [14]. To address this, the LoRA adapter [15] introduces an attention layer structure in the Transformer network, enabling fine-tuning of large models using less than 1% of the parameters, thereby making them applicable to downstream tasks. Motivated by these advancements, we propose the Oceannet model, which incorporates a multi-modal LoRA-tuning structure with spectrogram patch features. Through comprehensive experiments, we demonstrate that Oceannet achieves state-of-the-art performance in zero-shot learning and retrieval tasks.

---

[1] ChatGPT: https://chat.openai.com/
[2] Gemini: https://deepmind.google/technologies/gemini/

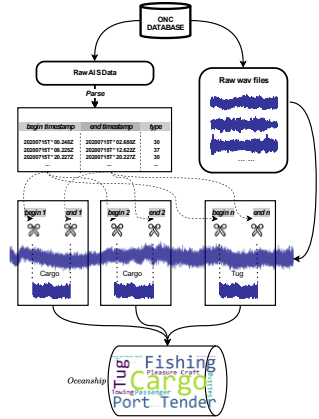 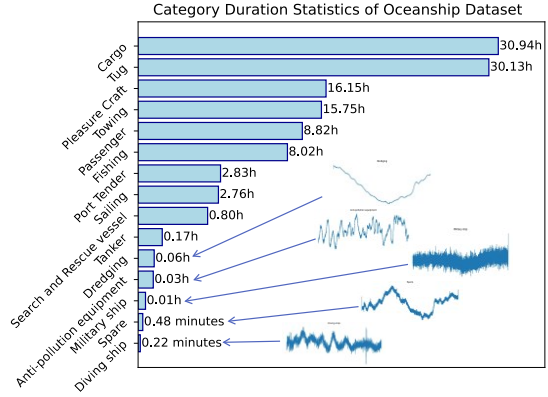

**Fig. 1.** provides a brief description of the data collection process for the Oceanship dataset.

**Fig. 2.** displays the duration distribution of different categories within the Oceanship dataset. It also showcases waveform graphs of sample audio from categories with short durations.

Our main contributions are as follows:

- We introduce Oceanship, the largest publicly available multi-labeled UATR dataset to date, comprising 107,540 audio-text pairs. This dataset strictly adheres to the AIS data processing protocol for accurate labeling of individual samples. Notably, 53,771 samples have undergone meticulous annotation, capturing fine-grained details such as location, heading, and speed.
- In our work, we redefine the supervised classification tasks in UATR and introduce two novel tasks: the 0-shot task and the Retrieval task. To the best of our knowledge, we are the first to conduct experiments on 0-shot learning and Retrieval in the UATR domain.
- We introduced Oceannet as a baseline model for generalizable UATR. To our knowledge, Oceannet is the pioneering patch-based multimodal learning framework in UATR domains. Additionally, we performed thorough experiments on zero-shot classification and retrieval tasks.

## 2  Dataset and Task Defination

### 2.1  Underwater Audio Target Recognition Dataset

As of present, two open-source underwater audio datasets are available: Deepship and ShipsEar. we will discuss the differences between the Oceanship dataset and existing UATR datasets, as well as delve into the details of these three datasets.

| Dataset Name | Category | Pairs | Duration(h) | AIS | Loc. | Heading | Spe. | MMSI |
|---|---|---|---|---|---|---|---|---|
| Shipsear | 11 | 91 | 3.13 | ✗ | ✓ | ✗ | ✓ | ✗ |
| Deepship | 4 | 465 | 47.22 | ✓ | ✗ | ✗ | ✗ | ✗ |
| Oceanship(CG.) | 15 | 53771 | 61.96 | ✓ | ✓ | ✗ | ✗ | ✓ |
| Oceanship(FG.) | 15 | 53769 | 59.21 | ✓ | ✓ | ✓ | ✓ | ✓ |
| Oceanship(Full) | **15** | **107540** | **121.17** | ✓ | ✓ | Half | Half | ✓ |

**Table 1.** Comparison of Oceanship with Deepship and Shipsear. The term "Category" denotes the assigned class number for each category within the dataset. "Duration" represents the cumulative length of all audio recordings in the dataset. "AIS" denotes adherence to the naming conventions outlined by the AIS protocol standards. "Loc." indicates Location, while "Spe." signifies Speed (cog and sog). "CG." refers to coarse-grained, "FG." refers to fine-grained.

**ShipsEar** [6], specifically, is a dataset comprising underwater recordings of ship and boat sounds during autumn 2012 and summer 2013 along various segments of the Spanish Atlantic coast in northwest Spain. It encompasses 90 recordings featuring 11 distinct vessel types and includes pertinent information such as channel depth, wind conditions, distance, and geographical coordinates. Moreover, it offers a notable advantage through its abundance of supplementary data, which can serve as valuable prior knowledge to enhance model training. However, it also presents a significant drawback: it comprises only 3 hours of audio data, posing a considerable challenge for a 11-classification task. In order to mitigate this limitation, the dataset provides two classification modes: 11-classification and 5-classification. To accommodate specific needs, the data can be grouped into 5 overarching categories based on the data subsets. Subsequent research primarily focuses on achieving 5-classification, while exploration of underwater audio classification for the 11-classification task remains limited.

**DeepShip** [4], established in 2021, is a benchmark dataset derived from ONC data, designed for underwater ship classification. It comprises 47 hours and 4 minutes of recordings featuring 265 different ships across four distinct classes. This dataset stands as the most authoritative resource for classification within its domain. Notably, the audio data extends fifteen times longer than that of the ShipsEar dataset, exhibiting a well-balanced distribution of instances across each category. This characteristic facilitates focused efforts on extracting audio spectrogram features and model design. Nonetheless, DeepShip has its limitations. Primarily, it encompasses a restricted number of categories, accounting for only 22% of communicative vessels documented in the AIS dataset categories. Hence, the generalization capability of the DeepShip dataset is limited.

**Oceanship**[3](Ours), as depicted in **Table 2**, our dataset was compared with two existing datasets, namely Deepship and Shipsear. Oceanship emerges as the largest dataset in terms of total duration, comprising 15 categories that cover 83% of the identifiable vessel names following the AIS naming protocol. Consequently, the Oceanship dataset encompasses all high-frequency vessels as stipulated by the AIS naming protocol. In terms of data volume, Oceanship exceeds the Deepship dataset by 2.5 times and the Shipsear dataset by 70 times in duration. Thus, the Oceanship dataset presents

---

[3] Oceanship: https://github.com/lizeyujack/oceanship

Oceanship: A Large-Scale Dataset for Underwater Audio Target Recognition 5

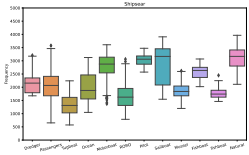
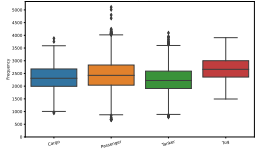
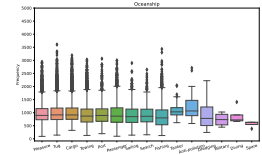

**Fig. 3.** Box plot of SE.  **Fig. 4.** Box plot of DS.  **Fig. 5.** Box plot of OS.

greater diversity in its content compared to the aforementioned datasets. The data is provided in three versions: the fine-grained version contains 53,771 audio samples with detailed annotations, the coarse-grained version comprises 53,769 audio samples lacking heading and speed information, and finally, the Full version includes 107,540 audio samples with coarse annotations. The sample size of the Oceanship (Full) dataset is at least 1,000 times larger than that of the ShipsEar dataset and over 200 times larger than that of the DeepShip dataset.

As illustrated in **Fig. 2**, the statistical analysis of the Oceanship categories is presented. Among these categories, Cargo and Tug contain the most abundant data, whereas Diving ship and Spare have the least amount of data. The observed data imbalance stems from the authors parsing nearly all vessels between July 15, 2020, and February 18, 2021. The duration of each category in the dataset corresponds to the frequency of occurrences within this timeframe.

All data utilized by Oceanship originates from ONC[4] (Ocean Networks Canada Society, 2017). The author obtains access to the ONC database through a token to download the encoded AIS communication messages and corresponding audio for the respective year. As illustrated in Figure 1a, the initial step involves decoding the AIS-encoded communication data to extract readable information. For instance, the original communication data appears as "34eG;tE000o:pUBKr00hf`s l0Drb,024", accompanied by the corresponding timestamp: 20200715T000000.036Z formatted as "YYYYMMDDTHHMMSS..z". Decoding this data entails parsing the AIS communication field, considering various attributes such as:{" **x** ": -123.4514 , " **y** ":48.7697 , " **sog** ":0.0 , "**cog**":18.6000 , "**true_heading**":285 , "**ais_timestamp**":"20200715T 000000.036 Z", "**mmsi**":"*********" , " **id** ":3}.

Given that the Maritime Mobile Service Identity (MMSI) acts as a distinctive identifier for vessels, a question mark is utilized to substitute the vessel code as needed. Subsequently, we retain the relevant information, particularly the timestamp details and ship type data. In the presented scenario, the vessel type remains indeterminate owing to a potential failure in MMSI query, rendering access to vessel-specific details, including vessel type data, unattainable.

We conducted frequency statistics on the three UATR datasets according to their categories and generated box plots for visualization. The boxplot of the Shipsear dataset is depicted in **Fig. 3**. The central tendency of the data exhibits significant variation, but there are almost no outliers, indicating a relatively stable data scenario.

---

[4] ONC DB: https://data.oceannetworks.ca/DatasetLandingPage?doi-dataset=10.21383/650d90face87-473c-b932-278519062ab5

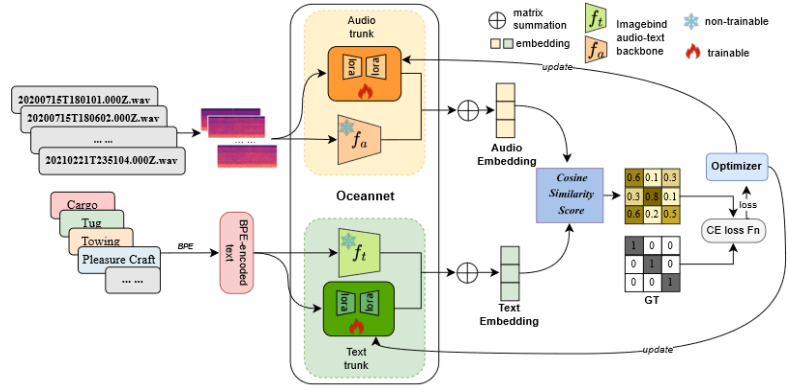

**Fig. 6.** The framework of Oceannet.

In **Fig. 4**, In contrast to Shipsear's performance, the central frequency point of Deepship data remains nearly the same within the range of 2000Hz-2500Hz. However, there are numerous outliers present, indicating a higher level of data diversity. As shown in **Fig. 5**, the Oceanship dataset exhibits the widest range of outliers, with the highest outlier frequencies exceeding three times the central frequency point. This wide range of frequencies can be attributed to the dynamic nature of the sampled underwater scenarios. The Oceanship dataset provides a richer variety of scenes, and the central frequency point at 1000Hz aligns well with the frequency range of underwater ship mechanical sonar.

| Model | Training set | Deepship Eval. | | | Shipsear Eval. | | |
|---|---|---|---|---|---|---|---|
| | | R@1 | R@3 | R@5 | R@1 | R@3 | R@5 |
| Imagebind | - | 4.79 | 18.02 | 33.56 | 1.29 | 5.68 | 11.09 |
| Imagebind(FT.) | Oceanship(Full) | 9.58 | 35.86 | 56.67 | 6.19 | 15.74 | 22.71 |
| Oceannet(ours) | Oceanship(CG.) | 36.72 | 55.03 | 70.28 | 13.73 | 15.79 | 31.58 |
| Oceannet(ours) | Oceanship(FG.) | 51.29 | 67.49 | 86.76 | 18.02 | 33.56 | 39.92 |
| Oceannet(ours) | Oceanship(Full) | **67.11** | **95.49** | **99.13** | **24.38** | **39.35** | **45.94** |

**Table 2.** The retrieval performance of audio-to-text UATR on the Deepship and Shipsear datasets is evaluated, where "FT." denotes fine-tuning, "CG." denotes coarse-grained, and "FG." denotes fine-grained.

### 2.2 Task Defination

We will present the historical supervised tasks in the domain of Underwater Audio Target Recognition (UATR) and put forward two novel tasks: zero-shot classification task and retrieval task.

**Supervised Classification Task in UATR**

While the classification task in UATR has been extensively discussed, there lacks a standardized description for supervised tasks. We define the training dataset as $S_{train} = \{x_i, y_i\}_{i=0}^{n_0}$ and the test dataset as $S_{test} = \{x_j, y_j\}_{j=0}^{m_0}$, where $\mathcal{P}_{\mathcal{XY}}^{train} = \mathcal{P}_{\mathcal{XY}}^{test}$. Both datasets originate from the same domain (derived from identical datasets), and typically, data from the same source exhibit consistent distributions. It is imperative that $\mathcal{Y}_{train} = \mathcal{Y}_{test}$. However, when data originates from distinct datasets, variations arise due to factors such as collection devices, locations, and wind speeds, resulting in dissimilar distributions: $\mathcal{Y}_{Deepship} \neq \mathcal{Y}_{Shipsear}$.

| Model | Feature Extractor | Testing Dataset | Deepship | | Shipsear | |
|---|---|---|---|---|---|---|
| | | Training Dataset | Oceanship (Full) | Shipsear | Oceanship (Full) | Shipsear |
| Efficientleaf[18] | MFCC | | 33.67 | 23.27 | 9.34 | 6.32 |
| Efficientleaf | TBN | | 37.11 | 28.63 | 18.36 | 13.12 |
| Efficientleaf | PCEN | | 35.78 | 26.09 | 12.49 | 2.13 |
| Underwater_snd[19] | CQT | | 43.36 | 23.71 | 25.66 | 7.12 |
| Underwater_snd | MFCC | | 40.88 | 25.25 | 21.83 | 6.60 |
| Underwater_snd | GFCC | | 32.97 | 18.55 | 19.72 | 6.98 |
| Oceannet(Ours) | MFCC,BPE | | **76.32** | **37.13** | **35.30** | **27.67** |

**Table 3.** Zero-shot performance evaluation of UATR on Deepship and Shipsear datasets.

**Zero-shot Task in UATR**

In this scenario, characterized by $\mathcal{P}_{\mathcal{XY}}^{train} \mid \mathcal{Y}_{DatasetA}$, $\mathcal{P}_{\mathcal{XY}}^{test} \mid \mathcal{Y}_{DatasetB}$, and $\mathcal{X}_{DatasetA} \cap \mathcal{X}_{DatasetB} = \emptyset$, it is termed a zero-shot situation. Leveraging the Oceanship dataset, we can conduct two varieties of zero-shot experiments. In the initial experiment, the Oceanship dataset serves as the training set: $\mathcal{P}_{\mathcal{XY}}^{train} \mid \mathcal{Y}_{Oceanship}$, while the Deepship dataset functions as the test set: $\mathcal{P}_{\mathcal{XY}}^{test} \mid \mathcal{Y}_{Deepship}$. Similarly, in the second experiment, the Oceanship dataset acts as the training set: $\mathcal{P}_{\mathcal{XY}}^{train} \mid \mathcal{Y}_{Oceanship}$, whereas the Shipsear dataset serves as the test set: $\mathcal{P}_{\mathcal{XY}}^{test} \mid \mathcal{Y}_{Shipsear}$.

**Retrieval Task in UATR**

In real-world coastal detection scenarios, the diversity of vessel types presents challenges for training a single network effectively. Therefore, we propose a multi-modal retrieval task aimed at underwater object recognition. This task leverages textual descriptions in the form of AIS vessel nomenclature. To facilitate the detection of vessel categories like Shipsear, we define a total of 26 text vessel query set: *{'Fishing', 'Motorboat', 'Port Tender', 'Spare', 'Trawler', 'Diving ship', 'Dredging', 'Towing', 'Search and Rescue vessel', 'Cargo', 'Pilot Vessel', 'Tanker', 'Pleasure Craft', 'Passenger', 'RORO', 'Sailboat', 'Military ship', 'Tug', 'Ocean liner', 'Mussel boat', 'Law Enforcement', 'Anti-pollution equipment', 'Medical Transport', 'Natural ambient noise', 'Sailing'}*. Textual data undergoes processing by a model to derive text embeddings, denoted as $E_q^t$. Moreover, embeddings for audio targets can also be generated, termed as $E_p^a$. Through the cosine similarity function, we identify the nearest text embedding $E_q^t$ from a set of M texts $E^t = \{E_1^t, \ldots, E_M^t\}$, determining the optimal match for the target vessel audio.

## 3  Model Architecture

Drawing inspiration from the partial success observed in transferring knowledge from CLIP to text-image classification, we opt to directly initialize our Oceannet with the complete Imagebind audio and text encoder. This approach aims to bolster its inherent cross-modal alignment capabilities. Additionally, during the pretraining phase, we utilize the Lora adapter to concurrently train both the audio and text modality branches while maintaining the parameters of the Imagebind backbone network frozen. As shown in **Fig. 6**, this framework facilitates feature extraction from both audio and text data, determining the correspondence between audio and text by evaluating the cosine similarity between the audio and text embeddings.

### 3.1   Text Encoder

For a given input text $T$, we utilize the Imagebind text encoder directly to derive the text representation, which is a modified version of ViT-Base[16]. Following the methodology of Imagebind, we utilize lower-cased byte pair encoding (BPE) with a vocabulary size of 49152 to tokenize the input text description. The text description is enclosed within [SOS] and [EOS] tokens to denote the beginning and end of the sequence. Subsequently, the tokenized text $f_{sos}^t, f_1^t, \ldots, f_{eos}^t$ is inputted into the transformer to explore correlations among each patch using masked self-attention.

### 3.2   Audio Encoder

We adhere to the methodology outlined in [13] for encoding audio. Specifically, a 5-second audio, sampled at 16kHz, is transformed into spectrograms using 128 mel-spectrogram (MFCC) bins. As spectrograms resemble 2D signals akin to images, we utilize a ViT model with a patch size of 16 and a stride of 10. Given the spectrogram $A_{\text{spec}} \in R^{H \times W \times C}$, we leverage a pre-trained Imagebind ViT model to derive the image embedding. The spectrogram processing involves initially dividing it into a sequence of $N = \frac{H \times W}{P^2}$ fixed-sized non-overlapping patches, where $P$ denotes the patch size. Subsequently, the patch sequence is mapped to 1D tokens $f_i^v|_{i=1}^N$ via a trainable linear projection.

### 3.3   Lora-Tuning

In the case of a pre-trained Imagebind weight matrix $W_0 \in R^{d \times k}$, its update is constrained through representation by a low-rank decomposition $W_0 + \Delta W = W_0 + BA$, where $B \in R^{d \times r}$, $A \in R^{r \times k}$, and the rank $r \ll \min(d, k)$. During training, the weights from $W_0$ remain frozen and do not receive gradient updates, while matrices $A$ and $B$ remain trainable. LoRA-tuning enables the indirect training of certain dense layers in a neural network by optimizing the rank decomposition matrices of the dense layers' changes during adaptation, while maintaining the pre-trained weights frozen, as illustrated in **Fig. 6**.

### 3.4 Loss Function

We utilize two encoders from the Imagebind audio trunk and text trunk to process the audio data input, denoted as $X_i^a$, and the text data input, denoted as $X_i^t$, separately. Here, $(X_i^a, X_i^t)$ represents an audio-text pair indexed by $i$. The audio encoder $f_{\text{audio}}(\cdot)$ with $LoRA_{audio}(\cdot)$ and the text encoder $f_{\text{text}}(\cdot)$ with $LoRA_{text}(\cdot)$ are employed to obtain the audio embedding, denoted as $E_i^a$, and the text embedding, denoted as $E_i^t$, respectively. These embeddings are further enhanced by projection layers in Equ.(1-2):

$$E_i^a = MLP_{audio}\big(f_{audio}(X_i^a) + LoRA_{audio}(X_i^a)\big) \quad (1)$$

$$E_i^t = MLP_{text}\big(f_{text}(X_i^t) + LoRA_{text}(X_i^t)\big) \quad (2)$$

The audio/text projection layer is a two-layer multilayer perceptron (MLP) with ReLU [17] as the activation function. It is used to map the encoder outputs into the same dimension $D$, denoted as $E_i^a$ and $E_i^t$ respectively, where $E_i^a, E_i^t \in R^D$. The model is trained using the contrastive learning paradigm, which compares the audio and text embeddings in pairs, as illustrated in Equ.(3):

$$L = \frac{1}{2N}\sum_{i=1}^{N}\left(\log\frac{\exp(E_i^a \cdot E_i^t/\tau)}{\sum_{j=1}^{N}\exp(E_i^a \cdot E_j^t/\tau)} + \log\frac{\exp(E_i^t \cdot E_i^a/\tau)}{\sum_{j=1}^{N}\exp(E_i^t \cdot E_j^a/\tau)}\right) \quad (3)$$

Where $\tau$ denotes a learnable temperature parameter used to scale the loss. The logarithmic terms consider either the audio-to-text. The variable $N$ typically represents the number of data instances. However, during the training phase, it is used as the batch size since it is computationally infeasible to compute the entire matrix of all data. Instead, we update the $LoRA_{audio}(\cdot)$ and $LoRA_{text}(\cdot)$ functions using batch gradient descent.

## 4 Experiments

### 4.1 Hyperparameters and Training Details

As described in Section 3, Oceanship serves as the training dataset for our model. Regarding the audio data, we employ a 5-second input duration, 240 hop size, 1024 window size, and 64 mel-bins for computing mel-spectrograms. Consequently, each input forwarded to the audio encoder has dimensions of (T=1024, F=64). For text data, we tokenize it with a maximum token length of 77. During training, we utilize the AdamW optimizer with $\beta_1 = 0.99$, $\beta_2 = 0.9$, and cosine learning rate decay with a base learning rate of 1e-5. The model undergoes training with a batch size of 64 on the Oceanship dataset for 200 epochs. The training as well as testing was performed on NVIDIA Tesla V100.

### 4.2   Audio-to-Text Retrieval

An experiment was conducted to compare the performance of the Imagebind model, fine-tuned and Lora-tuned, on an audio-to-text retrieval task. The results in Table 2 indicated that, unlike fine-tuning the 440 M Imagebind models for both audio and text modalities, Oceannet achieved state-of-the-art results when Lora-tuned on the Imagebind model, requiring training only 2.4 M parameters. This is because fine-tuning pretraining models with a large number of parameters poses significant challenges for convergence and computational burden.

### 4.3   Dataset Scale

Accordingly, we utilize Oceannet as our baseline model to perform comprehensive audio-to-text retrieval experiments, as outlined in Table 2. We adopt the metrics from to compute Recall values at various ranks in our task. The Oceanship (CG.) dataset comprises 53,769 text-audio pairs, while the Oceanship (FG.) dataset consists of 53,771 text-audio pairs, making them nearly equivalent in scale. The findings suggest that incorporating fine-grained labels into the text modality input enhances the retrieval performance of the model. While Oceannet has attained flawless performance on Deepship in the UATR retrieval task, it encounters challenges in demonstrating improvement in retrieval on the Shipsear validation set. One contributing factor is the variance between the Oceanship dataset and the Shipsear dataset, stemming from disparate data collection locations and sensor usage. Furthermore, there exists minimal overlap in category between the two datasets. Consequently, the R@1 scores of the aforementioned models on the Shipsear dataset, characterized by substantial domain distinctions, remain relatively low.

### 4.4   Zero-Shot Classification

To assess the model's generalization and the dataset's suitability, we conducted experiments detailed in Table 3. Evaluation was performed on two established UATR datasets, Deepship and Shipsear, utilizing top-1 Recall as the metric. Efficientleaf and Underwater_snd, established open-source models have attained state-of-the-art performance in supervised UATR tasks. In these models, alongside our proposed Oceannet, we conducted four sets of zero-shot classification experiments. These experiments entailed training on the Oceanship dataset and testing on the Deepship dataset; training on the Shipsear dataset and testing on the Deepship dataset; training on the Oceanship dataset and testing on the Shipsear dataset; and training on the Deepship dataset and testing on the Shipsear dataset. Consequently, our model, Oceannet, establishes new state-of-the-art benchmarks for zero-shot underwater audio classification across both Deepship and Shipsear datasets. Based on the aforementioned experimental findings, employing the Oceanship dataset as the pretraining dataset demonstrates superior generalization performance in zero-shot classification on the UATR dataset compared to utilizing the Deepship or Shipsear datasets. Thus, the Oceanship dataset demonstrates enhanced sample generalization compared to the two existing datasets.

## 5 Conclusion and Futurework

We address the generalization challenge in underwater audio classification tasks and establish definitions for supervised tasks, zero-shot classification tasks, and retrieval tasks within the underwater audio domain. To address the zero-shot classification and retrieval tasks, we introduce a baseline model called Oceannet, pretrained on the Oceanship dataset. Our dataset, Oceanship, is the largest publicly available multi-labeled UATR dataset to date, comprising 107,540 audio-text pairs. Compared to Deepship and Shipsear, not only is the annotation more refined, but also the variety of ship types is extensive, and it exhibits better generalization capabilities. Therefore, Oceanship is currently the most suitable pre-training dataset for UATR generalization tasks. Oceannet achieves a performance of 76.32% on the Deepship zero-shot classification task and 35.30% on the Shipsear zero-shot classification task. Future endeavors will entail gathering an even larger UATR dataset for training and exploring few-shot learning approaches in the testing set.

**Acknowledgments.** The data utilized in this study was acquired through the infrastructure provided by Ocean Networks Canada. This work was partially supported by China State Shipbuilding Corporation Limited- Shanghai Jiao Tong Univeristy Marine Equipment Foresight and Innovation Joint Fund, A Fast Online Learning Method for Deep Learning Networks for Hydroacoustic Target Recognition (4-B4) and the National Natural Science Foundation of China under Grant No.62301315.